# Transfer Learning-Based CNN Models for Plant Species Identification Using Leaf Venation Patterns

Bandita Bharadwaj[1], Ankur Mishra[2], Saurav Bharadwaj[3]

[1]School of Technology, Assam Don Bosco University, Guwahati, Assam, India

[2,3]Parul Institute of Engineering and Technology, Parul University, Vadodara, Gujarat, India

[3]Corresponding email: saurav.bharadwaj33162@paruluniversity.ac.in

**Abstract:** This study evaluates the efficacy of three deep learning architectures—ResNet50, MobileNetV2, and EfficientNetB0—for automated plant species classification based on leaf venation patterns, a critical morphological feature with high taxonomic relevance. Using the Swedish Leaf Dataset comprising images from 15 distinct species (75 images per species, totalling 1,125 images), the models were demonstrated using standard performance metrics during training and testing phases. ResNet50 achieved a training accuracy of 94.11% but exhibited overfitting, reflected by a reduced testing accuracy of 88.45% and an F1 score of 87.82%. MobileNetV2 demonstrated better generalization capabilities, attaining a testing accuracy of 93.34% and an F1 score of 93.23%, indicating its suitability for lightweight, real-time applications. EfficientNetB0 outperformed both models, achieving a testing accuracy of 94.67% with precision, recall, and F1 scores exceeding 94.6%, highlighting its robustness in venation-based classification. The findings underscore the potential of deep learning, particularly EfficientNetB0, in developing scalable and accurate tools for automated plant taxonomy using venation traits.

# 1. Introduction

Leaf venation, defined as the spatial arrangement of vascular bundles within the leaf blade, plays a vital role in plant physiology by facilitating the transport of water and nutrients, providing mechanical support, and exhibiting morphological traits that are often unique to specific species [1], [2]. Traditional plant identification techniques have predominantly focused on macroscopic characteristics such as leaf shape, margin, and colour [3]. While these methods can be effective, they are frequently time-consuming, subjective, and require considerable expertise in plant taxonomy [4]. Venation morphology offers a stable, species-specific, and morphometrically quantifiable trait that can be leveraged as a reliable biometric marker, enabling the development of automated, objective, and high-precision plant classification systems [5], [6].

Deep learning significantly enhances image-based plant identification by automating the robust extraction of discriminative features from raw images [7], [8]. Convolutional Neural Networks (CNNs) have demonstrated exceptional capability in capturing fine-grained venation patterns by learning hierarchical spatial features through multiple convolutional layers [9], [10]. CNN-based segmentation approach trained on more than 700 leaves from 50 Southeast Asian plant families achieved superior performance in extracting vein networks, attaining a precision-recall harmonic mean of 94.5% ± 6%, and enabling hierarchical loop decomposition to quantify multiscale statistics such as vein width, angles, and connectivity, thereby revealing network geometry [11]. Similarly, the D-Leaf model, incorporating CNN-based feature extraction and classification through multiple machine learning algorithms, achieved a high testing accuracy of 94.88%, outperforming traditional morphometric approaches based on Sobel-segmented veins, and demonstrated that CNN-extracted features integrated effectively with artificial neural network (ANN) classifiers [12]. Further advancements in multi-scale venation pattern analysis for medicinal plant recognition employed pre-processing techniques such as contrast enhancement and Frangi filtering, followed by three distinct architectures: a modified ResNet-50 adapted for venation-aware channels, a custom VenationNet optimized for hierarchical venation fusion, and a Dual-Stream CNN that independently processed texture and venation maps before attention-based fusion. While the modified ResNet-50 achieved a validation accuracy of 73.35%, the Dual-Stream CNN improved performance to 78% with a macro F1-score of 0.7485, and VenationNet delivered consistent results with a validation accuracy of 75.75%, underscoring the advantage of multi-scale venation-specific modelling in fine-grained plant species classification [13].

Leaf venation networks in angiosperms exhibit a typical hierarchical organization shared within clades, yet functional constraints such as hydraulic conductivity, transpiration efficiency, and tolerance to damage shape their morphospace distribution. Traditional quantification approaches have primarily relied on basic morphological metrics like vein length, diameter, branching angles, and areole area, which limit the ability to capture the full complexity of venation structures. Recent advancements have introduced network feature-based phenotyping workflows capable of high-throughput quantification by integrating deep neural network–based segmentation, undirected graph extraction, and topological feature computation, achieving classification accuracies of up to 90.6% across species and revealing one-dimensional morphospaces aligned with Pareto-optimal trade-offs between transport efficiency, construction cost, and damage tolerance [1]. Complementary to this, directional morphological filtering methods have been developed for automatic vein order classification in dense venation networks, enabling precise extraction of ecologically significant hierarchical traits with major vein deviations under 5 pixels and second-order completeness of 54.28% [5]. For rapid and scalable trait acquisition, object-oriented classification techniques adapted from remote sensing have demonstrated over 96% extraction accuracy across diverse leaf types and an 87.3% increase in vein density calculation speed compared to conventional approaches, using optimized multi-parameter thresholds for robust feature delineation [14]. Beyond geometric descriptors, topological phenotypes based on nested loop organization offer an orthogonal dimension in phenotypic space, enhancing species identification from leaf fragments and linking trait variability to developmental noise imprints on vascular architecture [15]. However, only a limited number of studies have focused on developing deep learning-based algorithms capable of classifying leaves based on venation patterns using portable camera-based systems for field applications, highlighting the need for further research in this area [16], [17].

This study investigates the application of three transfer learning-based CNN architectures—ResNet, EfficientNet, and MobileNet—for plant species classification using leaf venation patterns. ResNet incorporates residual connections that facilitate the training of deeper networks by mitigating the vanishing gradient problem, while EfficientNet adopts compound scaling to uniformly balance network depth, width, and resolution for improved efficiency. MobileNet employs depthwise separable convolutions to significantly reduce computational complexity, making it well-suited for deployment on resource-constrained devices. By focusing on venation, a morphologically distinctive feature, the study addresses the growing demand for accurate, automated, and scalable plant identification methods. A comparative evaluation of these architectures on a diverse dataset of leaf images yields valuable insights into their relative performance and suitability for venation-based classification tasks.

## 2. Materials and Methods

### 2.1 Dataset Description

Swedish Leaf Dataset, developed by researchers at Linköping University, serves as a standard benchmark for evaluating shape-based plant species classification algorithms. It comprises 1,125 colour images, with 75 samples from each of 15 distinct plant species, as illustrated in Fig. 1. Each image features a single leaf centrally positioned against a white background, thereby minimizing background noise and facilitating accurate shape analysis.

### 2.2 Pre-processing Pipeline

Sequential pre-processing steps are applied to a leaf image to enhance venation features for classification and analysis, as illustrated in Fig. 2. The raw image is initially acquired in RGB format using a camera, where each pixel contains intensity values for the red, green, and blue channels. This RGB image is subsequently converted into a grayscale image using a weighted sum of the RGB components. To suppress impulse noise—commonly referred to as salt-and-pepper noise—a two-dimensional median filter is applied. This filtering technique effectively smooths out isolated noisy pixels while preserving critical edge structures. The Sobel operator is employed to detect prominent edges in both horizontal and vertical directions, emphasizing intensity transitions corresponding to leaf boundaries and venation patterns. The directional gradient images generated by the Sobel filters are combined to produce a gradient magnitude image that enhances the visual contrast of vein structures against the background. The gradient image is complemented by inverting the pixel intensities, resulting in a transformation where dark veins on a bright background appear as bright veins on a dark background, thereby improving their visibility for further analysis.

### 2.3 Classification Models

Three deep learning models—ResNet50, MobileNetV2, and EfficientNetB0—were implemented using transfer learning. Each model employed a pre-trained CNN backbone originally trained on the ImageNet dataset, enabling the reuse of generalized feature representations learned from a large and diverse image corpus. The classification task was formulated as a supervised multi-class problem, wherein input images of leaves were labelled by species, and the objective was to predict the correct species based on venation characteristics.

#### 2.3.1 Transfer Learning Using ResNet50 Architecture

ResNet50 architecture is employed as a feature extractor. All leaf images were resized to 224×224 pixels to match the input requirements of the network. The dataset was managed using the ImageDataGenerator class, which performed channel-wise normalization using the dedicated pre-processing function for ResNet50. It was then split into 80% training and 20% validation subsets, with random shuffling applied to the training data. The base ResNet50 model was loaded without its top classification layers and initialized with pre-trained ImageNet weights. All convolutional layers were frozen to retain their learned representations. A custom classification head was appended to the base model, beginning with a global average pooling layer, followed by two dropout layers with rates of 0.5 and 0.3 to mitigate overfitting. Between these dropout layers, a fully connected dense layer with 256 neurons and ReLU activation was inserted. The final output layer was a dense layer with softmax activation, with the number of neurons corresponding to the number of target species. The model was compiled using the Adam optimizer and categorical cross-entropy loss. Training was conducted for up to 15 epochs, with early stopping enabled to monitor validation loss and restore the best-performing weights when no improvement was observed for five consecutive epochs [18].

**2.3.2 Transfer Learning Using MobileNetV2 Architecture**

MobileNetV2 is a lightweight CNN architecture optimized for efficiency on mobile and embedded systems. Input images were resized to 224×224 pixels, and a batch size of 32 was used. The ImageDataGenerator class normalized pixel values to the [0, 1] range by rescaling with a factor of 1/255. The dataset was split into 80% training and 20% validation sets and loaded using the flow_from_directory() method, which automatically inferred class labels from the folder structure. MobileNetV2 was initialized with include_top=False and preloaded with ImageNet weights. The convolutional base was frozen (trainable=False), and a custom classifier was appended. This classifier included a GlobalAveragePooling2D layer, followed by a dropout layer (rate = 0.5), a dense layer with 128 ReLU-activated units, and an additional dropout layer (rate = 0.3). A final dense output layer with softmax activation mapped the feature vector to class probabilities. The model was compiled using the Adam optimizer and categorical cross-entropy loss. An early stopping callback was employed to halt training if the validation loss did not improve over five consecutive epochs, with automatic restoration of the best weights. Training was conducted for a maximum of 30 epochs, and the history of training and validation metrics was stored for post-analysis [19].

**2.3.3 Transfer Learning Using EfficientNetB0 Architecture**

EfficientNetB0 architecture is an optimal trade-off between accuracy and model size. Input images were resized to 224×224 pixels, with a batch size of 32. Pre-processing was performed using the dedicated pre-processing function of EfficientNet, ensuring pixel scaling and normalization consistent with the original training setup. The dataset was partitioned into training and validation subsets using a validation split of 0.2 and loaded via flow_from_directory(), with appropriate shuffling applied to the training data. The EfficientNetB0 model was initialized with include_top=False and pre-trained ImageNet weights, and its base layers were initially frozen. A custom classification head was appended, comprising a global average pooling layer, dropout layers (with rates of 0.5 and 0.3), a dense layer with 256 ReLU-activated units, and a final softmax output layer for multi-class prediction. The model was compiled using the Adam optimizer and categorical cross-entropy loss, with accuracy as the evaluation metric. To prevent overfitting and enhance convergence, early stopping and a learning rate scheduler ReduceLROnPlateau were employed. The initial training phase lasted for 10 epochs, during which only the classification head was trainable. This was followed by a fine-tuning phase in which the final ten layers of the EfficientNetB0 base were unfrozen and trained alongside the classification head using the Stochastic Gradient Descent optimizer with a low learning rate and momentum. This strategy enabled selective adaptation of deeper features to the specific leaf venation dataset without compromising the generalization ability of the pre-trained network. Fine-tuning continued until the validation loss ceased to improve, as determined by the early stopping criterion [20].

**3. Results and Discussion**

**3.1 Performance Comparison and Confusion Matrix Analysis**

Table 1 provides a comprehensive comparative analysis of three deep learning models—ResNet50, MobileNetV2, and EfficientNetB0—applied to the task of plant species classification using leaf venation patterns. Model performance was evaluated using standard classification metrics, computed separately for training and testing datasets. The ResNet50 model achieved a training accuracy of 94.11%, indicating its strong capacity to learn discriminative features from the training set. However, a decline in testing accuracy to 88.45% points to overfitting, suggesting that the model may have memorized patterns in the training data without effectively generalizing to unseen samples. This discrepancy is further shown in its test F1-score of 87.82%, which, while moderately high, suggests an imbalance between precision and recall during inference on the test set. MobileNetV2 demonstrated improved performance across all metrics. It achieved a higher training accuracy of 96.34% and a corresponding testing accuracy of 93.34%, reflecting its ability to generalize more effectively. The model attained a testing precision of 94.32% and recall of 93.34%, resulting in an F1-score of 93.23%. These results indicate that

MobileNetV2 is capable of accurately identifying plant species but also maintains a stable trade-off between false positives and false negatives, making it suitable for practical classification scenarios. Among all the three evaluated models, EfficientNetB0 outperformed the others in both training and testing phases. It achieved a training accuracy of 95.45% and the highest testing accuracy of 94.67%. The model also reported consistent test precision (96.12%), recall (94.56%), and F1-score (95.23%), highlighting the ability to capture complex venation features and makes it a promising model for real-time plant identification applications.

Confusion matrices depicted in Fig. 3 provide a comprehensive visualization of classification performance of each model, facilitating an in-depth analysis of their capability to discriminate leaf venation patterns. In the case of ResNet50, while the diagonal is moderately pronounced—indicating a fair number of correct predictions—the substantial presence of off-diagonal elements reveals frequent misclassifications, as shown in Fig. 3(a). Species such as Ulmus carpinifolia and Fagus silvatica are often confused, implying that ResNet50 may struggle to capture the subtle morphological nuances in venation patterns required for fine-grained discrimination. MobileNetV2 model shows a more coherent structure, with a sharper diagonal and reduced off-diagonal noise, reflecting better inter-class separability and improved overall accuracy, as shown in Fig. 3(b). Confusion persists among closely related species, suggesting that although MobileNetV2 offers enhanced feature extraction, its discriminatory power remains challenged by fine-scale venation similarities. Compared to the above two models, the confusion matrix for EfficientNetB0 displays a dominant diagonal with minimal off-diagonal dispersion, indicating the highest classification fidelity among the three models, as shown in Fig. 3(c). Sparse misclassification patterns, even for morphologically similar species like Ulmus carpinifolia, underscore the robustness of EfficientNetB0 in learning intricate venation features, making it particularly effective for precise species-level classification.

**3.2 Receiver Operating Characteristic Curves**

Receiver operating characteristic (ROC) curves for the three models used for plant species classification based on leaf venation patterns are shown in Fig. 4. Each subplot illustrates the per-class ROC curves for the respective model, offering insights into its ability to discriminate among the plant species. In Fig. 4(a), the ResNet50 model demonstrates strong overall classification performance, with area under the curve (AUC) values predominantly ranging from 0.97 to 1.00. Although this indicates better classification model, the slightly reduced AUC for Ulmus glabra (0.97) suggests a relatively lower performance in identifying this particular species, potentially due to overlapping venation features with other taxa. Fig. 4(b) illustrates the performance of the MobileNetV2 model, which exhibits improved ROC characteristics compared to ResNet50. The majority of classes achieve near-perfect AUC values of 1, demonstrating the heightened ability of the model to differentiate between species. Minor deviations are observed for Ulmus carpinifolia, Ulmus glabra and Betula pubescens, three classes with AUC scores of 0.99, reflecting a marginal decline in classification certainty for these three species. EfficientNetB0, as depicted in Fig. 4(c), outperforms both preceding models in terms of ROC metrics, achieving AUC values of 0.98 or higher for all plant species. Most classes reach a perfect AUC of 1.00, and even the least performing class, Ulmus glabra attains an AUC of 0.98.

The Fig. 5 illustrates the mean ROC curves for the three models employed in the classification of plant species using leaf venation patterns. ROC curves, which plot the true positive rate (sensitivity) against the false positive rate (1-specificity), provide a comprehensive evaluation of the discriminative power of each model across all classification thresholds. ResNet50 exhibits a high mean AUC of 0.9967, indicating strong predictive performance, although it is marginally lower than that of the other two models. MobileNetV2 and EfficientNetB0 achieve higher mean AUC values of 0.9983 and 0.9984, respectively, reflecting their enhanced ability to maintain a good balance between sensitivity and specificity.

**3.3 Precision-Recall (PR) Curves**

Multiclass Precision-Recall (PR) curves for the three transfer learning models used for plant species classification based on leaf venation patterns are shown in Fig. 6. These curves illustrate the trade-off between precision and recall for each plant species, providing insights into the model behavior under class imbalance conditions. As shown in Fig. 6(a), the ResNet50 model demonstrates moderate to high prediction capabilities, with AUC values ranging from 0.80 to 1.00. Lower AUC values are observed for Ulmus glabra (0.80), Ulmus carpinifolia (0.83), and Fagus silvatica (0.87), indicating reduced classification confidence for these species. The majority of classes achieve AUC values close to or equal to 1.00, reflecting strong model performance across most categories. MobileNetV2, shown in Fig. 6(b), demonstrates improved performance over ResNet50, with most species achieving AUC values above 0.90. Several species—including Quercus, Alnus incana, Populus tremula, and Tilia—attain perfect AUC scores of 1.00. AUC values for previously underperforming species, such as Ulmus glabra and Fagus silvatica, improve to 0.91 and 0.96, respectively, indicating a more balanced precision–recall relationship and a reduced risk of misclassification. EfficientNetB0, as shown in Fig. 6(c), delivers the most robust classification performance. Nearly all species achieve AUC values of 0.97 or higher, with perfect scores (AUC = 1.00) for a substantial number of classes. While Ulmus glabra maintains a slightly lower AUC of 0.83 and Fagus silvatica achieves 0.97, these values still represent high predictive reliability. EfficientNetB0 outperforms the other models in maintaining a strong balance between precision and recall.

Fig. 7 presents the mean PR curves of the three models used for plant species identification based on leaf venation features. PR curves are particularly valuable for evaluating classification performance on datasets with class imbalance, as they emphasize the ability of the model to correctly identify positive instances. The ResNet50 model demonstrates strong performance with an AUC of 0.96, although it shows a slightly steeper decline in precision at higher recall levels (Fig. 7(a)). However, both MobileNetV2 and EfficientNetB0 exhibit flatter PR curves with higher AUCs of 0.98, indicating a more balanced trade-off between precision and recall (Fig. 7(b) and 7(c)). These results suggest that while all three models are effective, MobileNetV2 and EfficientNetB0 offer marginally better performance in minimizing false positives while maintaining high recall across species.

**3.4 Score Plots**

Fig. 8 summarizes the per-class evaluation results for three transfer learning models applied to plant species classification using leaf venation patterns. Each subplot presents class-wise bar plots comparing four key performance metrics. ResNet50 model demonstrates uneven performance across species. While species such as Quercus, Populus tremula, and Tilia exhibit high scores across all metrics, other classes—including Ulmus carpinifolia, Ulmus glabra, and Fagus silvatica—show lower values, particularly in F1-score and recall (Fig. 8(a)). MobileNetV2 model exhibits more consistent performance across most species. Precision and recall are generally well-balanced, resulting in high F1-scores for nearly all classes. Although some variation remains, previously underperforming species such as Ulmus glabra and Fagus silvatica demonstrate moderate improvements, reflecting a better balance between correct identification and missed instances (Fig. 8(b)). EfficientNetB0 achieves the most stable and robust performance among the three models, with the majority of species attaining high scores across all four metrics. Even the lower-performing species, such as Ulmus glabra and Fagus silvatica, show marked improvements in accuracy and precision compared to the other models (Fig. 8(c)).

## 3.5 Learning Curves

Fig. 9 illustrates the training and validation accuracy, along with training and validation loss, over successive epochs for the three deep learning models employed in this study. Fig. 9(a) and 9(b) correspond to the ResNet50 model. As shown, the training accuracy increases with each epoch, while the validation accuracy begins to plateau around epoch 10. This trend indicates that the model effectively learns from the training data but reaches a saturation point early, suggesting limited capacity to further improve on unseen data. The associated loss curves exhibit a steady decline, though a persistent gap between training and validation loss is evident. This discrepancy hints at mild overfitting, where the model performs better on training data compared to the validation set. Fig. 9(c) and 9(d) depict the learning behaviour of the MobileNetV2 model. This model demonstrates rapid convergence, achieving high validation accuracy within the first few epochs. The close alignment between training and validation accuracy throughout the training process indicates strong generalization and minimal overfitting. The training and validation loss curves show a sharp initial decrease, followed by smooth stabilization, which is indicative of efficient learning dynamics and model robustness. Fig. 9(e) and 9(f) represent the performance of the EfficientNetB0 model. Among the three architectures, EfficientNetB0 exhibits the highest and most stable validation accuracy, with minimal fluctuations across epochs. The corresponding loss curves converge and maintain low values throughout training, reinforcing the effective learning of the model. This behaviour highlights superior performance of EfficientNetB0, followed by MobileNetV2. ResNet50 lags in both convergence speed and validation performance.

Compared to prior studies that focused on specialized venation-specific architectures or CNN-based segmentation pipelines for feature extraction, our research adopts a direct classification approach using three widely recognized deep learning models—ResNet50, MobileNetV2, and EfficientNetB0—trained on leaf venation images. Earlier works, such as the CNN-based vein network extraction achieving a 94.5% precision-recall harmonic mean [11] or the D-Leaf model with 94.88% accuracy through hybrid CNN–ANN integration [12], demonstrated strong performance but often relied on intermediate vein segmentation and handcrafted post-processing for trait quantification. Similarly, multi-scale venation models like VenationNet and Dual-Stream CNN [13] achieved moderate accuracies (73.35–78%) while emphasizing domain-specific feature fusion rather than generalizable architectures. However, this study shows that MobileNetV2 (testing accuracy 93.34%, F1-score 93.23%) and especially EfficientNetB0 (testing accuracy 94.67%, F1-score 95.23%) can achieve comparable or superior accuracy without complex segmentation or custom venation-aware layers, suggesting that modern lightweight and compound-scaled CNN architectures are capable of directly learning venation-discriminative representations from raw images. Moreover, while traditional venation phenotyping studies often optimize for trait extraction accuracy or morphospace mapping [1], [4], this approach prioritizes end-to-end species classification performance, positioning it as a more deployment-ready solution for field and portable camera-based applications where pre-processing may be limited.

## 4. Conclusion and Future Scopes

This study systematically evaluated the performance of three advanced deep learning architectures—ResNet50, MobileNetV2, and EfficientNetB0—for plant species classification using leaf venation patterns. Utilizing the Swedish Leaf Dataset comprising 15 species, the models were rigorously assessed across various performance metrics. Among the evaluated models, ResNet50 exhibited signs of overfitting, with relatively lower testing accuracy and F1 score despite its strong training performance. MobileNetV2 achieved a favourable trade-off between accuracy and computational efficiency, suggesting its suitability for lightweight and real-time field-based applications. EfficientNetB0 emerged as the most robust model, delivering the highest classification accuracy of 94.67%, thereby underscoring its superior generalization capabilities.

While the current findings affirm the efficacy of deep neural networks for venation-based plant identification, several directions remain open for future exploration. Expanding the dataset to include a wider variety of species and incorporating diverse environmental conditions—such as changes in lighting, leaf orientation, or background clutter—could enhance model robustness. Combining venation patterns with other morphological features such as shape, texture, and colour may further improve classification performance. Developing optimized, mobile-compatible frameworks would enable practical deployment of these models in real-time field-based applications, thereby supporting on-the-go plant identification for farmers.

**Statements & Declarations:**

**Funding:** The authors declare that no funds, grants, or other support were received during the preparation of this manuscript.

**Competing Interests:** The authors have no relevant financial or non-financial interests to disclose.

**Authors Contributions**: All authors contributed to the conception and design of the study. Data pre-processing and analysis were performed by B.B. and A.M. The first draft of the manuscript was written by S.B., and all authors commented on previous versions. All authors read and approved the final manuscript.

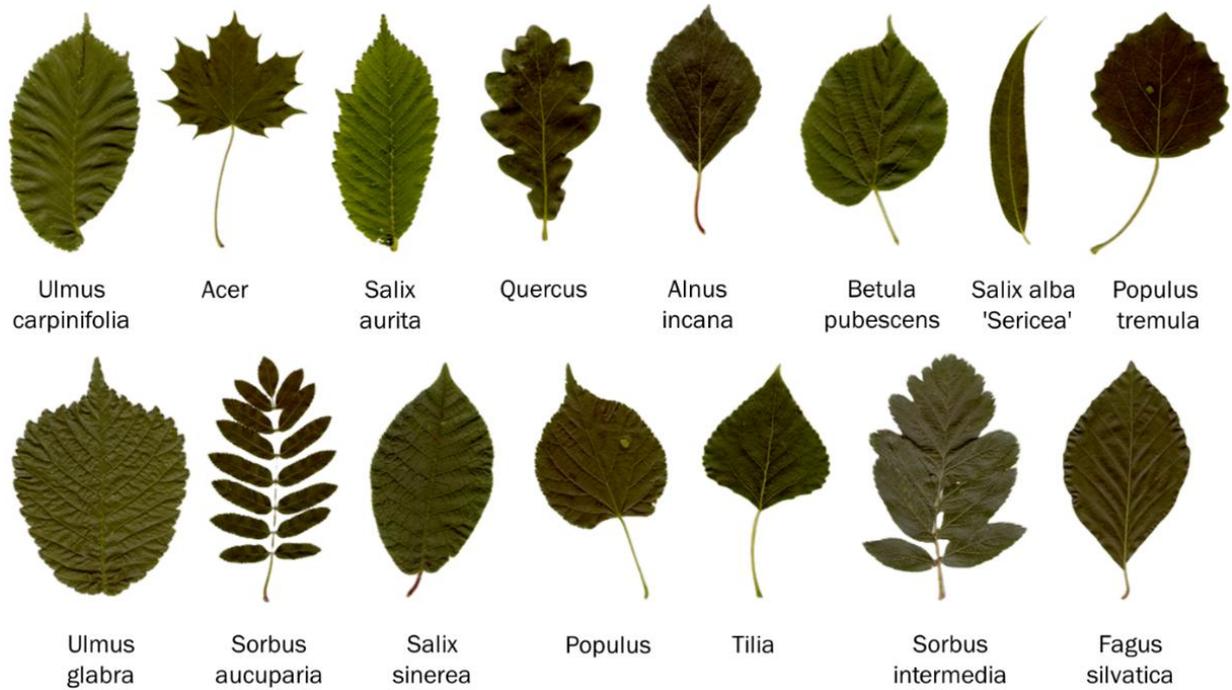

Fig. 1. Sample leaves of fifteen different plant species used in the study.

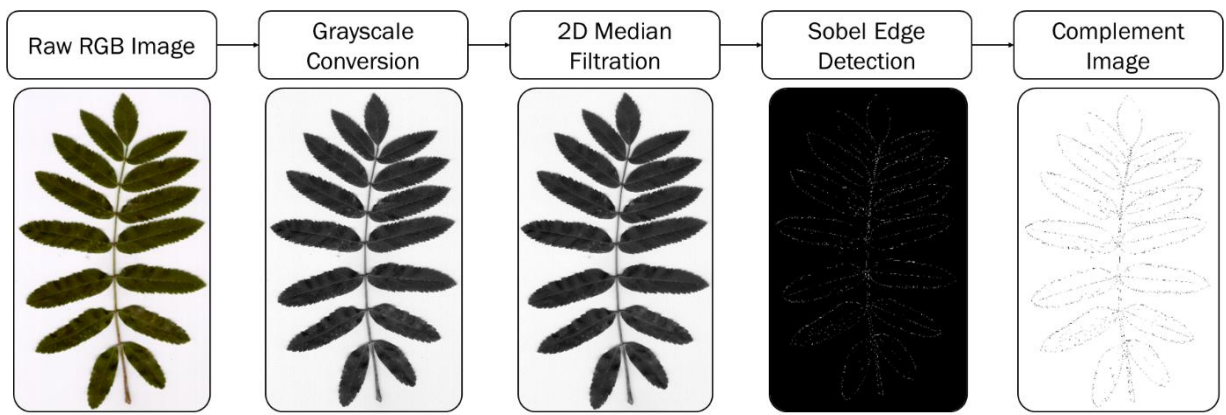

Fig. 2. Sequential image preprocessing pipeline for venation-based leaf analysis. The process begins with raw RGB image acquisition, followed by grayscale conversion, median filtering, Sobel edge detection to extract venation patterns, and image complementation to enhance venation features for downstream analysis.

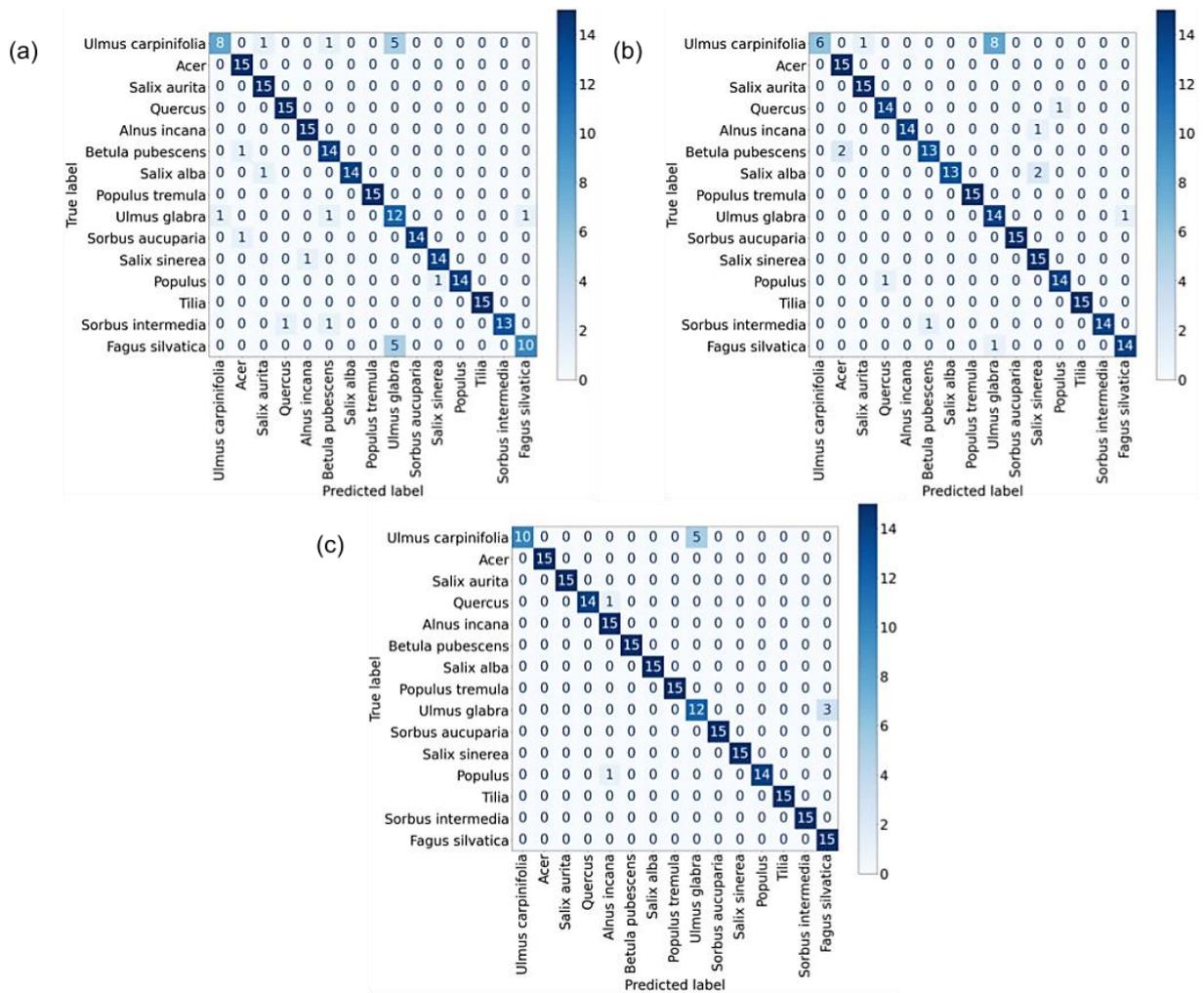

Fig. 3 Confusion matrices for plant species classification using transfer learning-based models: (a) Model 1 (ResNet50), (b) Model 2 (MobileNetV2), and (c) Model 3 (EfficientNetB0). Each matrix shows the classification performance for each species, with the rows representing true labels and the columns representing predicted labels. Darker shades on the diagonal indicate higher correct classification counts.

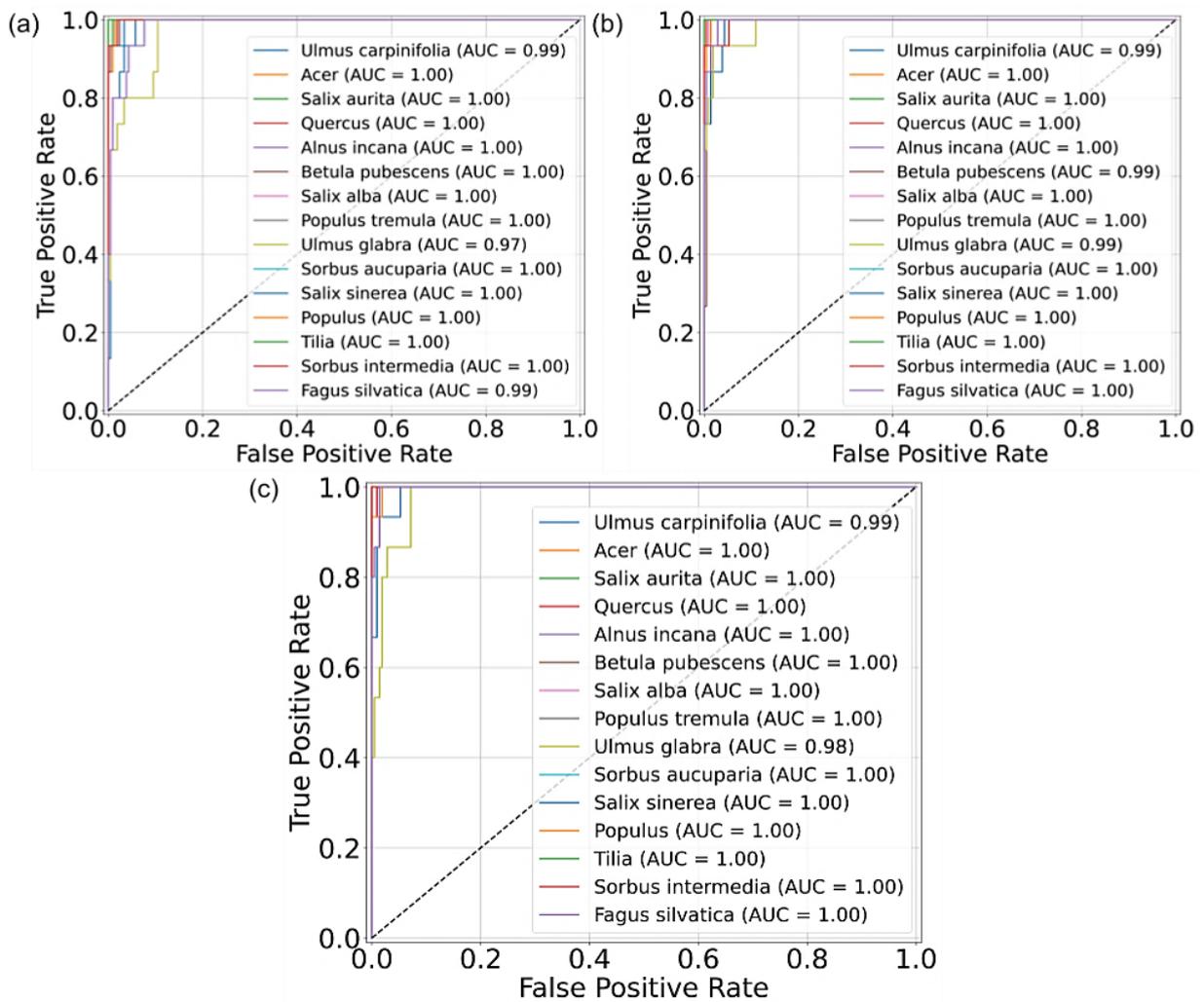

Fig. 4 ROC curves for plant species classification using transfer learning-based models: (a) Model 1 (ResNet50), (b) Model 2 (MobileNetV2), and (c) Model 3 (EfficientNetB0).

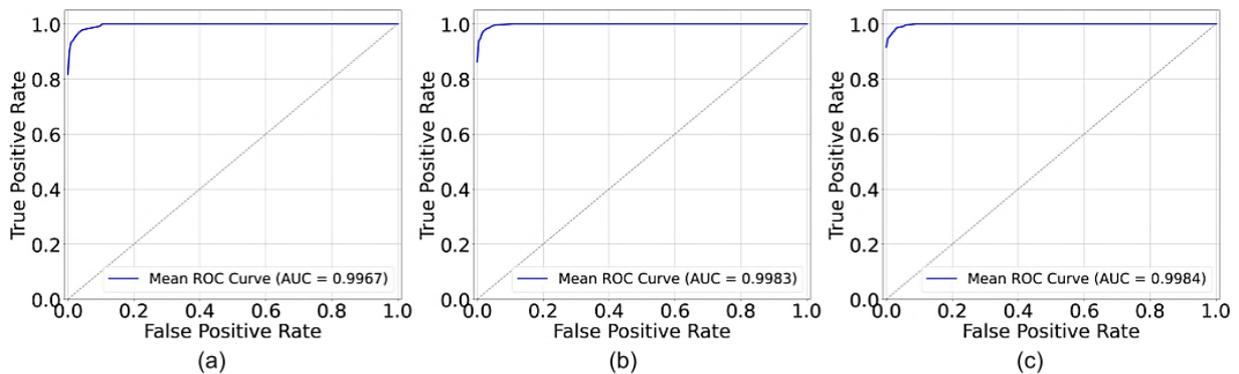

Fig. 5 Mean ROC curves for plant species classification using transfer learning-based models: (a) Model 1 (ResNet50), (b) Model 2 (MobileNetV2), and (c) Model 3 (EfficientNetB0).

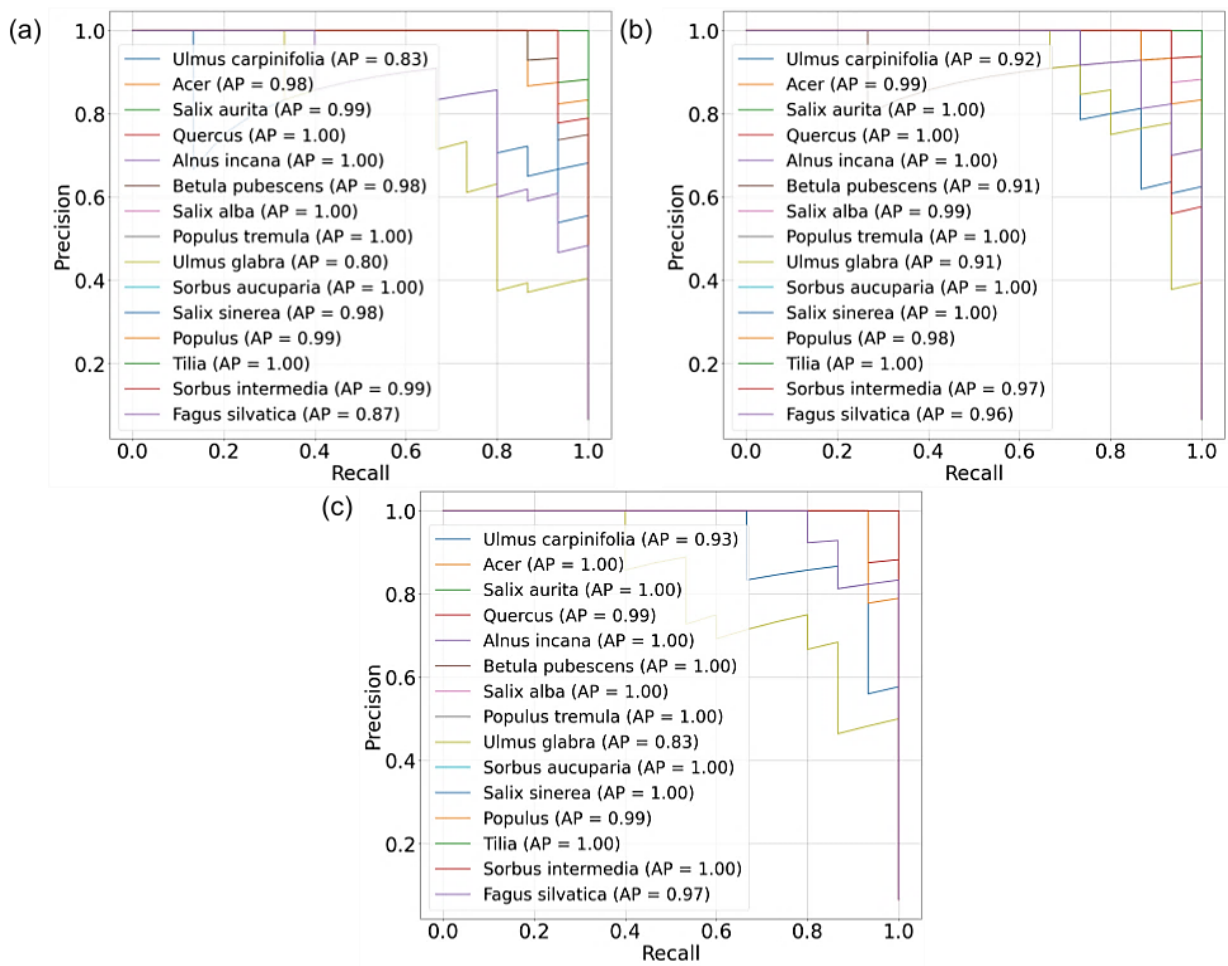

Fig. 6 PR curves for plant species classification using transfer learning-based models: (a) Model 1 (ResNet50), (b) Model 2 (MobileNetV2), and (c) Model 3 (EfficientNetB0).

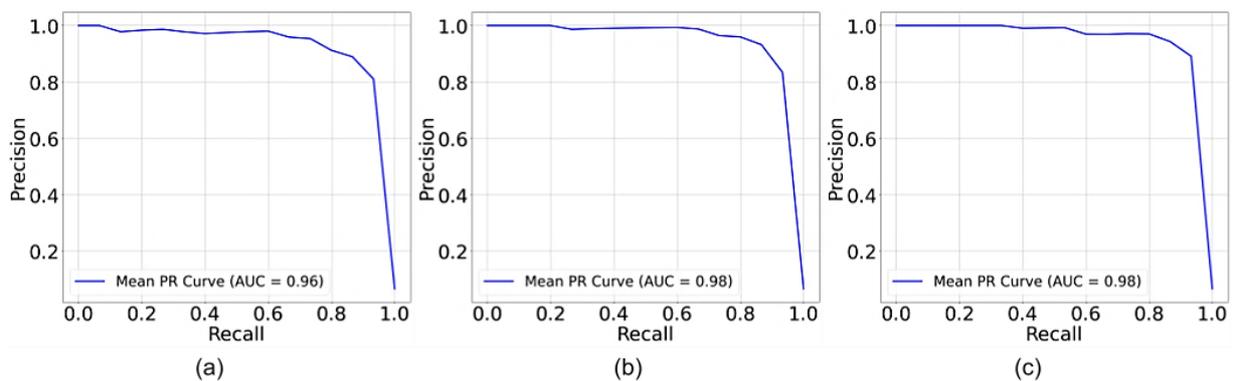

Fig. 7 Mean PR curves for plant species classification using transfer learning-based models: (a) Model 1 (ResNet50), (b) Model 2 (MobileNetV2), and (c) Model 3 (EfficientNetB0).

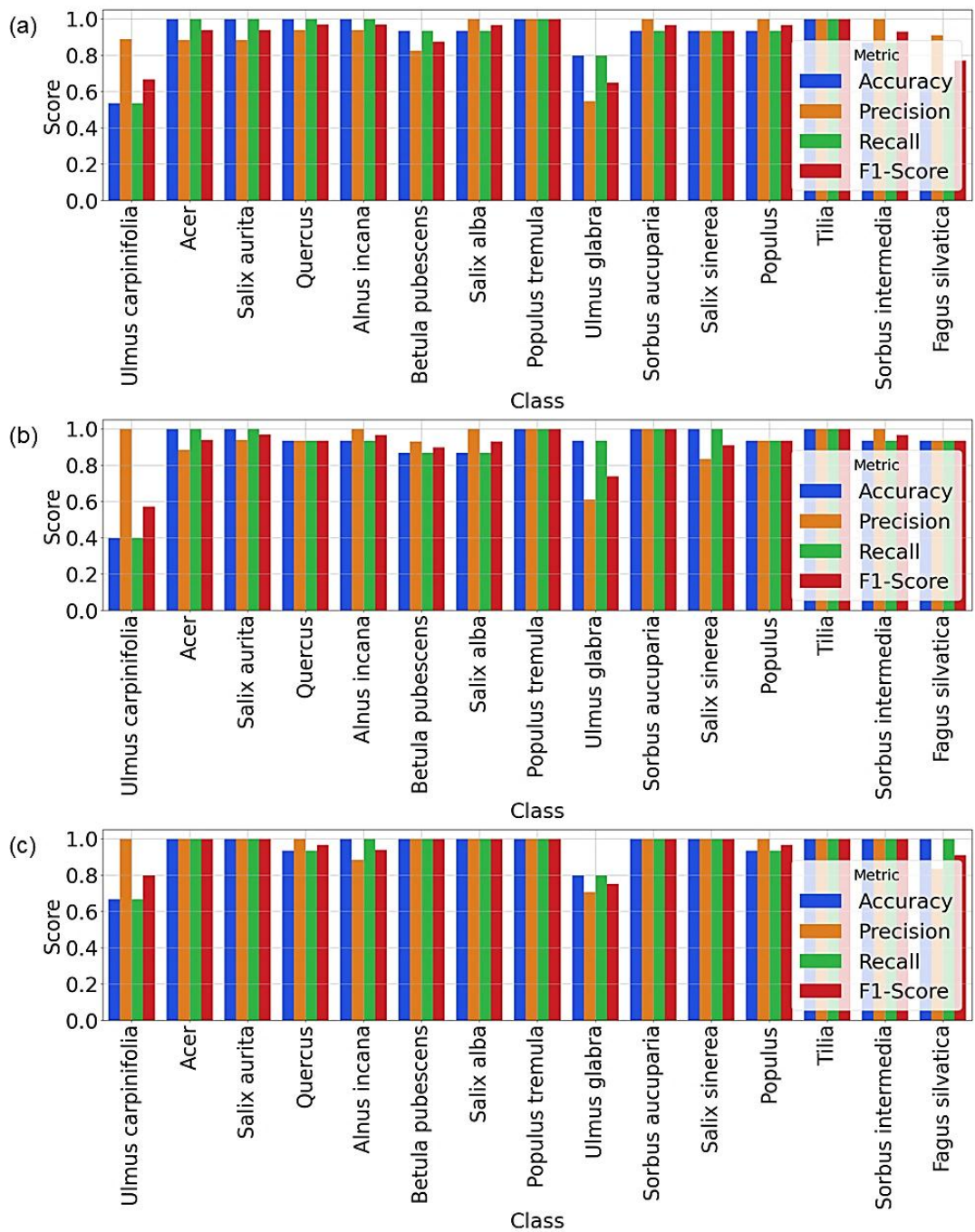

Fig. 8 Per-class evaluation metrics for plant species classification using transfer learning-based models: (a) Model 1 (ResNet50), (b) Model 2 (MobileNetV2), and (c) Model 3 (EfficientNetB0).

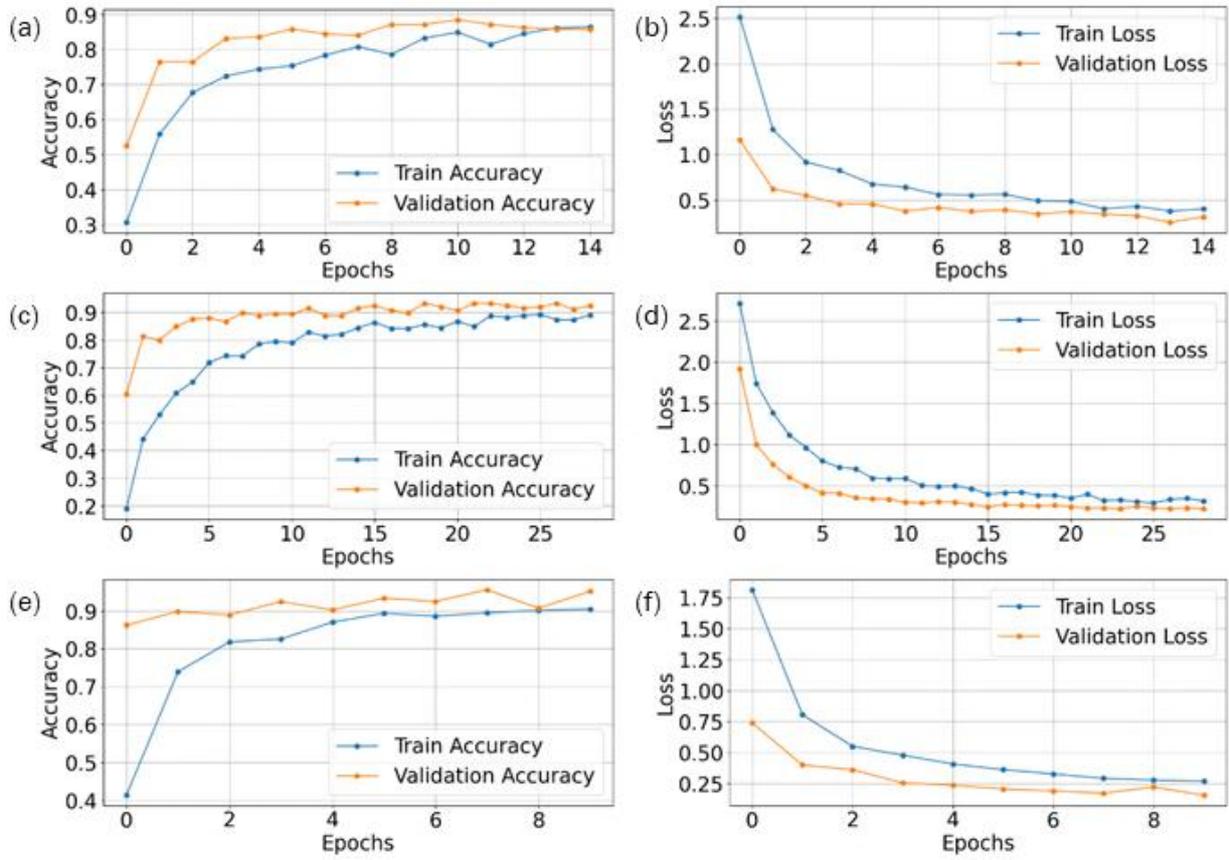

Fig. 9 Training and validation accuracy and loss curves for plant species classification based on leaf venation patterns using three models: (a, b) Model 1 (ResNet50), (c, d) Model 2 (MobileNetV2), and (e, f) Model 3 (EfficientNetB0).

Table 1: Performance comparison of different classification models.

| Models | Training Metrics | | | | Testing Metrics | | | |
|---|---|---|---|---|---|---|---|---|
| | Accuracy | Precision | F1 Score | Recall | Accuracy | Precision | F1 Score | Recall |
| Model 1 | 0.9411 | 0.9427 | 0.9411 | 0.9412 | 0.8845 | 0.8921 | 0.8782 | 0.8845 |
| Model 2 | 0.9634 | 0.9653 | 0.9634 | 0.9634 | 0.9334 | 0.9432 | 0.9323 | 0.9334 |
| Model 3 | 0.9545 | 0.9609 | 0.9552 | 0.9545 | 0.9467 | 0.9613 | 0.9465 | 0.9467 |